\documentclass[runningheads]{llncs}

\usepackage{eccv}

\usepackage{eccvabbrv}

\usepackage{graphicx}
\usepackage{booktabs}
\usepackage[misc]{ifsym}
\usepackage{amsmath}
\usepackage{amssymb}
\usepackage{tabularx}
\usepackage{multirow}
\usepackage{bbding}
\usepackage{amssymb}
\usepackage{pifont}
\usepackage{bbm}
\usepackage[dvipsnames]{xcolor}

\definecolor{cvprblue}{rgb}{0.21,0.49,0.74}
\definecolor{frenchblue}{rgb}{0.0, 0.45, 0.73}
\usepackage{colortbl}
\DeclareMathOperator*{\argmax}{arg\,max}  % Note 

\makeatletter
\def\blfootnote{\xdef\@thefnmark{}\@footnotetext}
\makeatother

\usepackage[accsupp]{axessibility}  % Improves PDF readability for those with disabilities.

\usepackage{hyperref}

\usepackage{orcidlink}

\begin{document}

% ---------------------------------------------------------------
% TODO REVIEW: Replace with your title
\title{Bridging Synthetic and Real Worlds for Pre-training Scene Text Detectors} 

% TODO REVIEW: If the paper title is too long for the running head, you can set
% an abbreviated paper title here. If not, comment out.
\titlerunning{FreeReal for Pre-training STDs}

% TODO FINAL: Replace with your author list. 
% Include the authors' OCRID for the camera-ready version, if at all possible.
\author{Tongkun Guan\inst{1}\and
Wei Shen\inst{1}\textsuperscript{(\Letter)}\and
Xue Yang\inst{2}\and \\ Xuehui Wang\inst{1}\and
Xiaokang Yang\inst{1}
}

% TODO FINAL: Replace with an abbreviated list of authors.
\authorrunning{T.~Guan et al.}
% First names are abbreviated in the running head.
% If there are more than two authors, 'et al.' is used.

% TODO FINAL: Replace with your institution list.
\institute{MoE Key Lab of Artificial Intelligence, AI Institute, Shanghai Jiao Tong University \email{\{gtk0615,wei.shen\}@sjtu.edu.cn} \and
Shanghai AI Laboratory\\
\email{yangxue@pjlab.org.cn}}

\maketitle
\blfootnote{\noindent \textsuperscript{\Letter}Corresponding author.}

\begin{abstract}
Existing scene text detection methods typically rely on extensive real data for training. 
Due to the lack of annotated real images, recent works have attempted to exploit large-scale labeled synthetic data (LSD) for pre-training text detectors. However, a synth-to-real domain gap emerges, further limiting the performance of text detectors. 
Differently, in this work, we propose FreeReal, a real-domain-aligned pre-training paradigm that enables the complementary strengths of both LSD and unlabeled real data (URD).
Specifically, to bridge real and synthetic worlds for pre-training, a glyph-based mixing mechanism (GlyphMix) is tailored for text images.
GlyphMix delineates the character structures of synthetic images and embeds them as graffiti-like units onto real images. Without introducing real domain drift, GlyphMix freely yields real-world images with partial annotations derived from synthetic labels. Furthermore, when given free fine-grained synthetic labels, GlyphMix can effectively bridge the linguistic domain gap stemming from English-dominated LSD to URD in various languages. Without bells and whistles, FreeReal achieves average gains of 1.97\%, 3.90\%, 3.85\%, and 4.56\% in improving the performance of FCENet, PSENet, PANet, and DBNet methods, respectively, consistently outperforming previous pre-training methods by a substantial margin across four public datasets. Code will be available at \url{https://github.com/SJTU-DeepVisionLab/FreeReal}.
\keywords{Scene text detection \and Pre-training paradigm \and Domain-bridging method}
\end{abstract}    
\section{Introduction}
\label{sec:intro}

\begin{figure}[t]
  \centering
  \includegraphics[width=0.8\linewidth]{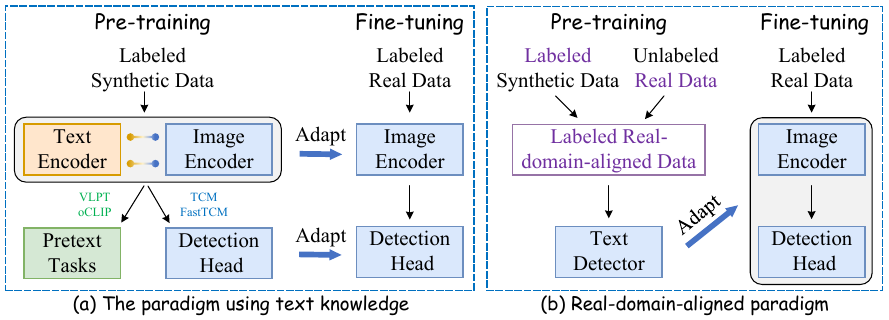}
  \caption{We present different pre-training paradigms for enhancing text detectors. 
  % (a) represents the existing pre-training paradigm by leveraging visual and textual knowledge. (b) denotes our paradigm by leveraging the intrinsic qualities of real-world unlabeled images with annotations derived from synthetic labels. 
  % Experiments demonstrate that our method achieves comprehensive performance improvements on public datasets.
  }
  \label{fig:0}
\end{figure}
Scene text detection aims to predict the bounding box or polygon for each arbitrarily shaped text instance within an image. 
This technology has diverse applications, including online education, instant translation, and product search, owing to the prevalence of text-laden images.
However, despite their considerable progress, supervised methods~\cite{DB,guan2022industrial,FCENet,PSENet,wang2020contournet} still struggle with an increasing demand for real data with labor-intensive annotations, especially when detecting complicated text images. 

In contrast, recent research efforts~\cite{SKTM,VLPT,oCLIP} have focused on pre-training scene text detectors using large-scale labeled synthetic data (LSD), which leverage both visual and textual knowledge to extract text feature representations, as illustrated in Fig.\,\ref{fig:0}(a). 
For example, VLPT~\cite{VLPT} formulates three cross-modality pretext tasks to enhance the performance of the backbone, and oCLIP~\cite{oCLIP} focuses on aligning visual and textual information to assist text detectors. 
Nevertheless, these well-curated designs are limited to using synthetic images due to their reliance on labeled word transcriptions. This limitation will lead to a noticeable synth-to-real domain gap when models are fine-tuned on real data, negatively impacting text detection performance. Therefore, exploring the potential of unlabeled real data (URD) is of great importance as they are readily available. However, the pre-training on URD has seldom been studied. One of the few works~\cite{MTM} attempts to pre-train text detectors on URD via masked image modeling~\cite{MIM} (step 1) and on LSD via supervised learning (step 2), which will re-transfers pre-trained weights from the real domain to the synthetic domain.

Differently, our objective is to jointly leverage LSD and URD by bridging their gap for enhancing the text detector pre-training, as shown in Fig.\,\ref{fig:0}(b).
The primary challenge is how to effectively modulate these URD and LSD without encountering real domain drift. 
In light of this, we initially investigate the adaptation capabilities of existing domain-bridging methods~\cite{mixup,cutmix,classmix}. This involves training a classifier to distinguish between URD and LSD and subsequently assessing the confidence of these domain-bridging methods in aligning with the real domain. As shown in Fig.\,\ref{fig:1}(b), their highest score is only 78.9\%. The observation implies that these domain-bridging methods may introduce intermediate domains and disrupt the real domain.

To conquer this domain drift issue, we propose FreeReal, a real-domain-aligned pre-training paradigm, leveraging synthetic labels to guide text detector learning on real-domain-aligned images.
Specifically, a glyph-based mixing mechanism, named GlyphMix, is tailored for text images to create real-world images with annotations derived from synthetic labels. This involves delineating the glyphs of synthetic images based on their internal structure and embedding them as graffiti-like units onto any unlabeled real images. GlyphMix offers several advantages: 1) extracting glyph structures in a cost-effective way without introducing extra boundary priors; 
2) leveraging synthetic labels for free while greatly minimizing real domain drift (achieving a confidence score of 95.7\%).

As a byproduct, FreeReal can bridge the language-to-language gap, when adapting features learned from English-dominated LSD to URD in various languages. Specifically, given free character-level synthetic labels, FreeReal enables the pre-trained model to easily learn character features, since characters are the underlying unit of any language. Subsequently, during fine-tuning, these learned character features are adaptively combined according to the rules (\emph{e.g.,} word- and sentence-level) provided by available benchmarks.

Building upon these insights, we break the trend of recent pre-training methods for scene text detectors (PSTD) that combine increasingly complex pretext task designs or more training modules to improve performance.
Without bells and whistles, we focus on exploring the potential of integrating the LSD and URD. 
Remarkably, despite using 90\% less LSD than recent SOTAs, our method still attains a 1.2\% gain, effectively underscoring the superiority of our pre-training paradigm. 
As shown in Table \ref{tb:expr1}, FreeReal consistently surpasses recent SOTAs by a large margin.
The main contributions are summarized as follows:
\begin{figure*}[t]
  \centering
  \includegraphics[width=0.8\linewidth]{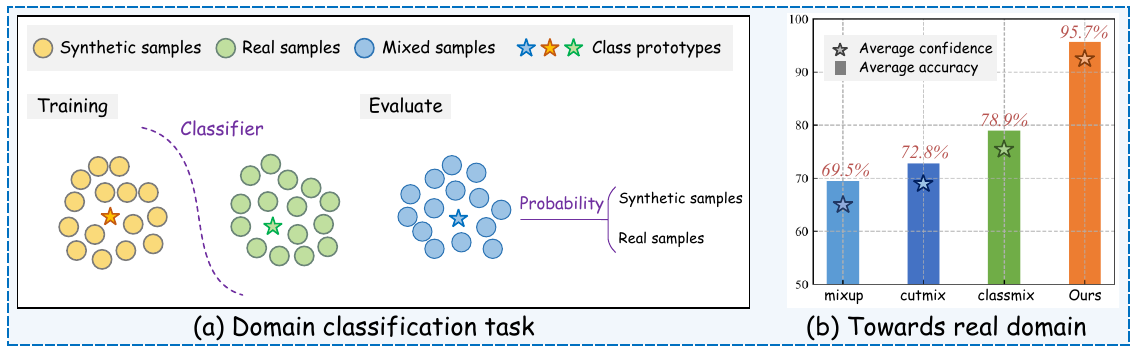}
  \caption{We present the domain classification task and the performances of different domain bridging ways between LSD and URD. In the text detection area, experiments demonstrate that our domain bridging method significantly performs better in aligning with the real domain. This is easy to explain: (I) Mixup may lead to unexpected pixel-wise ambiguities; (II) CutMix may cause the absence of many text regions and incomplete semantics of texts; (III) ClassMix will introduce salient boundary priors; (IV) GlyphMix (ours) effectively preserves the semantic information of text without causing real domain drift and bringing extra boundary priors. 
  }
  \label{fig:1}
\end{figure*}
\begin{itemize}
  \setlength{\itemsep}{0.0pt}
  \setlength{\parsep}{0pt}
  \setlength{\parskip}{0pt}
  \item We are the first to simultaneously integrate labeled synthetic data and unlabeled real data for enhancing the pre-training of text detectors. FreeReal shows its superiority and significant performance gains over other complex alternatives, serving as a simple yet strong baseline for future PSTD studies.
  \item 
  We effectively address the challenges of synthetic-to-real and language-to-language domain adaptation.
  Without bringing boundary priors and causing real domain drift, we freely harness the strengths of real-world text images with annotations derived from synthetic labels.
\end{itemize}

% \begin{figure}[t]
% \centering
% \includegraphics[width=0.9\linewidth]{./graph/char_word_sentence_plus.pdf}
% \caption{
% We illustrate differences in character, word, and sentence spacing between LSD and URD. When leveraging their inherent properties to improve text detection, the divergence in spacing can confuse the pre-trained model, making it challenging to interpret text layout and semantic context. For example, the model predictions cover text areas well but yield low F-measure values in (c). Green areas are filled by ground-truths, while red polygons indicate network predictions. 
% }
% \label{fig:2}
% \end{figure}

\section{Related Work}
\label{sec:formatting}

\noindent \textbf{Scene Text Detection.}
Existing STD methods delineate text regions by regression-based or segmentation-based representations. 
The regression-based methods~\cite{zhou2017east,FCENet,liu2020abcnet,guan2022industrial,wang2019arbitrary, MOST,qin2021mask} directly regress the bounding points to enclose text regions, and gradually increase the number of points as required to detect complex scenes, \emph{e.g.}, arbitrary texts. 
The segmentation-based methods regard texts of arbitrary shapes by grouping pixel-level~\cite{PAN,DB++,DB,sun2023feature}, character-level~\cite{CARFT,SIGA}, segment-level~\cite{long2018textsnake,SegLink} or contour-level~\cite{wang2020textray,wang2020contournet,qin2021fc} segmentation results by post-processing. 

\noindent \textbf{Pre-training for Scene Text Detector.}
In early research, \cite{gupta2016synthetic,long2020unrealtext,liao2020synthtext3d} proposed to pre-train scene text detectors by generating synthetic data. These data engine tools follow the same paradigm: they first stack multiple off-the-shelf models to perform semantic segmentation and depth estimation analysis on background images, and then blend pre-processed texts into these images. However, the multi-stage, pre-generated methods incur large computational costs (GPU and CPU resources) and substantial latency, making them unsuitable for online training tasks where immediate data processing is essential.
Inspired by recent advancements in pre-training techniques~\cite{chen2020simple,caron2021emerging,radford2021learning,CCD}, the integration of text and visual cross-modal information into scene text detectors has garnered increasing attention. 
One representative work involves enhancing the backbone with text transcriptions. 
For example, STKM~\cite{SKTM} proposes a pioneering work that pre-trains the backbone through an image-level text recognition task. 
VLPT~\cite{VLPT} formulates three cross-modality pretext tasks to boost better text representations derived from the backbone. 
Another representative work is to endow the entire detector with pre-trained vision-text knowledge. Inherited from CLIP~\cite{radford2021learning}, TCM incorporates the Image and Text encoders into the text detector to refine text regions using cross-modal visual-language priors. A persistent challenge still remains—the synth-to-real domain gap limits the detection performance. 

\noindent \textbf{Semi-Supervised Learning.}
Due to the scarcity of expensive annotations, semi-supervised learning (SSL) methods are becoming increasingly important. These methods extract knowledge from a reduced amount of annotated images in a supervised manner, and
from a larger amount of unlabeled images in an unsupervised way.
% In scenarios where both labeled and unlabeled data are from the same domain, the challenge involves leveraging the intrinsic qualities of the unlabeled data to improve model performance. 
% In scenarios where labeled and unlabeled data belong to different domains, the challenge involves adapting a model trained on labeled data from a source domain to perform well on unlabeled data from a target domain. This situation is called unsupervised domain adaptation.
Commonly, a self-training framework for SSL is employed, which encourages a model (student) to learn unlabeled images from pseudo-labels generated by the up-to-date optimized model (teacher). To avoid overfitting incorrect pseudo-labels, recent advancements have focused on pseudo-label refinement by improving the reliablity~\cite{Flexmatch,Fixmatch} and pseudo-label regularization by cross-view consistency~\cite{xu2022semi,laine2016temporal,tarvainen2017mean}. 
Specifically, some domain adaption works~\cite{zhan2019ga,wu2020synthetic,chen2021self} for STD focus on minimizing domain discrepancies in the feature space, including self-training and adversarial alignment methods. The self-training methods adopt the source data to pre-train a model for generating pseudo labels on target data, then use them to retrain the model. Adversarial alignment approaches aim to align the feature distribution and learn the domain-invariant features between the source and target domains. Recently, they have been replaced by the student-teacher framework, which significantly reduces domain gaps through dual-branch knowledge transfer~\cite{yang2022survey}.
\begin{figure}[t]
  \centering
  \includegraphics[width=1.0\linewidth]{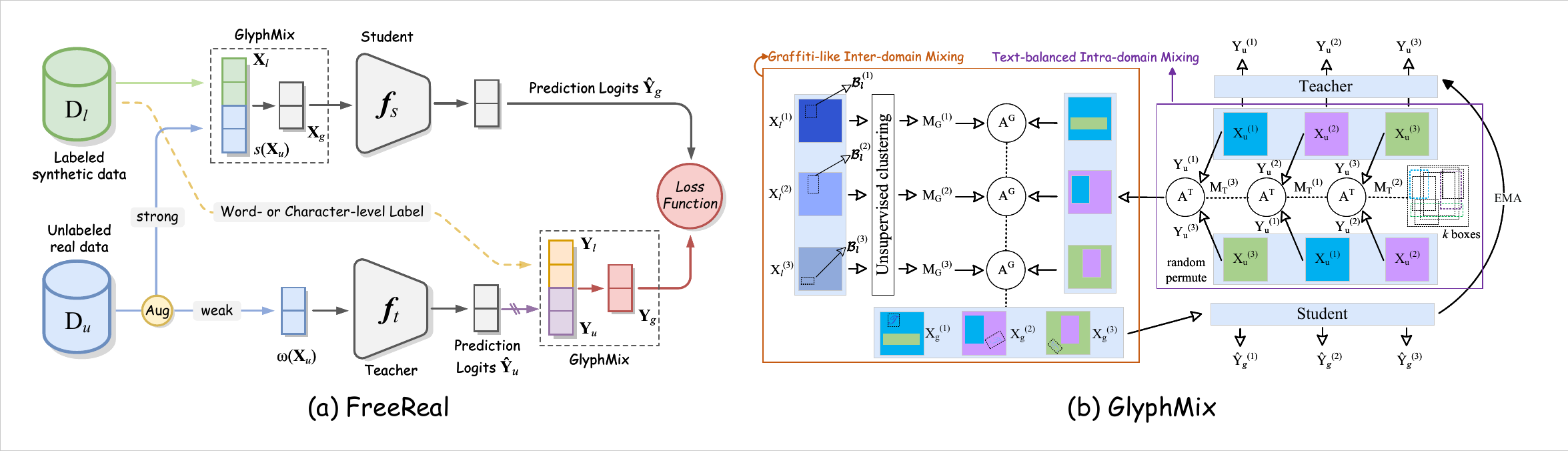}
  \caption{Our network pipeline in a student-teacher framework.
  }
  \label{fig:3}
\end{figure}
\section{Methodology}

\noindent \textbf{Overview.} 
Given a labeled synthetic set $\mathcal{D}_{l} = {\{(\mathbf{X}_{l}^{(i)},\mathcal{B}_{l}^{(i)})\}_{i=1}^{N_{l}}}$ with $N_{l}$ samples, $\mathbf{X}_{l}^{(i)} \in \mathbb{R}^{H\times W\times 3}$ represents an image and $\mathcal{B}_{l}^{(i)}=\{(p_{lt}^{i},p_{lb}^{i},p_{rt}^{i},p_{rb}^{i})\}_{i=1}^{n}$ is its corresponding set of bounding boxes. 
$p$ refers to the values of the vertice along the abscissa and ordinate, and $n$ denotes the number of bounding boxes. $H$ and $W$ are the height and width of the image. Similarly, the unlabeled real set is denoted as $\mathcal{D}_{u} = {\{\mathbf{X}_{u}^{(i)}\}_{i=1}^{N_{u}}}$ with $N_{u}$ samples. 
Our objective is to jointly leverage labeled synthetic data $\mathcal{D}_{l}$ and unlabeled real data $\mathcal{D}_{u}$ by bridging their gap for enhancing the pre-training of scene text detectors.

The student-teacher framework exhibits substantial efficacy due to its dual capacity for knowledge transfer between its two branches, which is typically employed to mitigate the domain gap (e.g., data and multimodal). Accordingly, we customize it for aligning real and synthetic text data, as illustrated in Fig.\,\ref{fig:3}.
For notation simplification in the following introduction, we omit the superscript $i$ as mentioned above.

Specifically, the segmentation-based text detector DBNet~\cite{DB} is employed as our baseline model, primarily due to its impressive real-time performance, versatility, and practical applicability. 
In the teacher branch, the teacher model parameterized by $\mathbf{f}_{t}$, is responsible for generating truncated-gradient logits $\mathbf{\hat{Y}}_{u}\in \mathbb{R}^{H\times W}$ from a weakly augmented unlabeled real image ${w(\mathbf{X}_{u})}$. The pseudo label $\mathbf{Y}_{u}\in\{0,1\}^{H\times W}$ is then obtained by binarizing the logit map $\mathbf{\hat{Y}}_{u}$.
In the student branch, the process begins by generating a pixel-wise binary map $\mathbf{Y}_{l}\in\{0,1\}^{H\times W}$, where 1 is assigned to locations within all bounding boxes from annotation $\mathcal{B}_{l}$, and 0 otherwise. The labeled synthetic image $\mathbf{X}_{l}$ and the strongly augmented unlabeled real image $s(\mathbf{X}_{u})$ are combined to create $\mathbf{X}_{g} = \texttt{A}(s(\mathbf{X}_{u}), \mathbf{X}_{l})$, with $\texttt{A}(\cdot)$ denoting our proposed GlyphMix mechanism, which will be introduced later. Subsequently, the mixed image $\mathbf{X}_{g}$ serves as input to train the student model, under the supervision of mixed pseudo-labels $\mathbf{Y}_{g} = \texttt{A}(\mathbf{Y}_{u}, \mathbf{Y}_{l})$, by minimizing loss in the following:
{\setlength\abovedisplayskip{2pt}
\setlength\belowdisplayskip{2pt}
\begin{equation}
\begin{aligned}
  \mathcal{L} &= \ell_{DB} (\mathbf{\hat{Y}}_{g}, \mathbf{Y}_{g}, \mathbf{E}),
  \label{eq2}
\end{aligned}
\end{equation}
}

\noindent where $\ell_{DB}$ denotes the original loss function provided by DBNet~\cite{DB} and $\mathbf{\hat{Y}}_{g}$ represents the prediction logits generated by the student model. 
The matrix $\mathbf{E}\in \mathbb{R}^{H\times W}$, which indicates positive regions for loss calculation, is set as an all-one matrix by default.

To filter out the unreliable predictions in mixed pseudo-labels $\mathbf{Y}_{g}$, we adaptively adjust the matrix $\mathbf{E}$ during pre-training.
Specifically, for the labeled synthetic image $\mathbf{X}_{l}$, the positive regions $\mathbf{E}_{l}$ are an all-one matrix.
For the unlabeled real image $\mathbf{X}_{u}$, we compute the pixel-wise entropy map on $\mathbf{\hat{Y}}_{u}$ and then apply a dynamic entropy threshold $\zeta$ to obtain its positive regions $\mathbf{E}_{u}$ as follows:
{\setlength\abovedisplayskip{2pt}
\setlength\belowdisplayskip{2pt}
\begin{equation}
\begin{aligned}
  \mathbf{E}_{u} = \mathbb{I}(-\mathbf{\hat{Y}}_{u} \log \mathbf{\hat{Y}}_{u}\leq \zeta),
\end{aligned}
\end{equation}
}

\noindent where $\mathbb{I}(\cdot)$ denotes the indicator function. 
The matrix is subsequently formulated as $\mathbf{E} = \texttt{A}(\mathbf{E}_{u}, \mathbf{E}_{l})$. In our experiments, the threshold $\zeta$ is determined by the value corresponding to the top $\gamma$ percent of the entropy map. As the reliability of pseudo-label gradually improves, we smoothly update $\gamma$ using \texttt{np.linspace(80, 20, t)}, where \texttt{t} denotes the total number of training iterations. 

Finally, the teacher model $\mathbf{f}_{t}$ gradually updates itself via the exponential moving averaging (EMA) of the student weights $\mathbf{f}_{s}$ as follows:
{\setlength\abovedisplayskip{2pt}
\setlength\belowdisplayskip{2pt}
\begin{equation}
  \begin{aligned}
    \mathbf{f}_{t} = \alpha \mathbf{f}_{t} + (1-\alpha) \mathbf{f}_{s},
  \end{aligned}
\end{equation}
}

\noindent where $\alpha$ is a momentum parameter, set as 0.996 by default.

\subsection{GlyphMix}
Although some efforts~\cite{mixup,cutmix,classmix} attempt to aggregate the advantages of synthetic and unlabeled images through versatile transformations, they inevitably introduce intermediate domains, leading to real domain drift. 
In this section, we present a domain-bridging way tailored for text images, allowing for the free use of synthetic labels while ensuring consistent alignment with the real domain.

\noindent \textbf{1) Graffiti-like Inter-domain Mixing.} 
Given an input synthetic image $\mathbf{X}_{l}$, its corresponding bounding box annotations $\mathcal{B}_{l}$ and converted pixel-wise binary map $\mathbf{Y}_{l}$, our goal is to perform character segmentation. This process identifies all character structure regions and produces a mask $\mathbf{M}_{G}$.

To achieve this, we begin with an unsupervised clustering task based on the grayscale values of pixels, avoiding the introduction of extra pixel-level annotation costs. 
This involves aligning and cropping the image according to the coordinates of each bounding box, yielding a set of image patches $\mathcal{T} =\{\mathbf{T}^{(i)} | \mathbf{T}^{(i)} \in \mathbb{R}^{w_{i}\times h_{i}\times 3}\}_{i=1}^{n}$.
Subsequently, within each image patch $\mathbf{T}^{(i)}$, we cluster the pixels into two regions: the glyph region representing character structures and the background region. 
For simplicity, we focus our study on K-means to implement the clustering task, but other clustering approaches with predefined categories can be used. 
Following this, the glyph regions from all image patches are pasted to an all-zero image to create the character structure mask $\mathbf{M}_{G}$.

Given the pseudo label $\mathbf{Y}_{u}$ produced by the teacher model on an unlabeled real image $\mathbf{X}_{u}$, we use $\mathbf{M}_{G}$ to generate the inter-domain mixing image and its corresponding label. This process can be formulated as follows:
{\setlength\abovedisplayskip{2pt}
\setlength\belowdisplayskip{2pt}
\begin{equation}
  \begin{cases}
  \begin{aligned}
    & \texttt{A}^{G}(\mathbf{X}_{u},\mathbf{X}_{l}) \leftarrow \mathbf{M}_{G} \odot \mathbf{X}_{l} + (\textbf{1}-\mathbf{M}_{G}) \odot \mathbf{X}_{u},\\
    & \texttt{A}^{G}(\mathbf{Y}_{u},\mathbf{Y}_{l}) \leftarrow \mathbf{M}_{G} \odot \mathbf{Y}_{l} + (\textbf{1}-\mathbf{M}_{G}) \odot \mathbf{Y}_{u},
  \end{aligned}
  \end{cases}
\end{equation}
}

\noindent where \textbf{1} is an all-one matrix and $\odot$ denotes the element-wise multiplication operation.
$\mathbf{M}_{G}$ indicates pixels occupied by the glyph need to be copied from the synth domain and pasted to the real domain. Thus the mixed image closely aligns with the real domain, taking advantage of the freely available synthetic labels. Put simply, this process produces real-world images with partial annotations derived from synthetic labels.

\noindent \textbf{2) Text-balanced Intra-domain Mixing.}
Real-world text images inherently exhibit complexity, and the collected unlabeled data often struggles to comprehensively represent the full spectrum of real-world scenarios. Within a student-teacher framework, intra-domain mixing serves as an effective perturbation technique, enriching the diversity of real data and enhancing feature consistency and robustness.
However, when randomly generating and applying binary masks to construct intra-domain mixing images and their corresponding labels, there is a risk of missing many text regions, especially considering that text regions typically occupy only a small fraction of an image. To mitigate this challenge, our objective is to retain more text regions while still maintaining randomness. 

In light of this, we initiate the process by randomly selecting two unlabeled real images, denoted as $(\mathbf{X}_{u}^{(1)}, \mathbf{X}_{u}^{(2)})$, and obtaining the corresponding segmentation pseudo-labels, $(\mathbf{Y}_{u}^{(1)}, \mathbf{Y}_{u}^{(2)})$, produced by the teacher model. 
Subsequently, $k$ rectangular boxes $\{(p_{lt}^{i},p_{lb}^{i},p_{rt}^{i},p_{rb}^{i})\}_{i=1}^{k}$ are randomly generated to crop the pseudo-label $\mathbf{Y}_{u}^{(2)}$ based on their coordinates. This operation yields a set of local pseudo-label regions $\mathcal{S} =\{\mathbf{S}^{(i)}|\mathbf{S}^{(i)} \in \mathbb{R}^{{w}_{i}\times {h}_{i}}\}_{i=1}^{k}$. 
Following this, we identify the rectangular box that contains the maximum text regions by calculating $\argmax_{i} \sum_{j=1}^{{w}_{i}\times {h}_{i}}\mathbf{S}^{(i)}_{j}$, and convert it into a binary mask, denoted as $\mathbf{M}_{T}$. 
Finally, we employ $\mathbf{M}_{T}$ to generate the intra-domain mixed image and its corresponding label, following the formulated procedure:
{\setlength\abovedisplayskip{2pt}
\setlength\belowdisplayskip{2pt}
\begin{equation}
  \begin{cases}
  \begin{aligned}
    & \texttt{A}^{T}(\mathbf{X}_{u}) \leftarrow \mathbf{M}_{T} \odot \mathbf{X}_{u}^{(2)} + (\textbf{1}-\mathbf{M}_{T}) \odot \mathbf{X}_{u}^{(1)},\\
    & \texttt{A}^{T}(\mathbf{Y}_{u}) \leftarrow \mathbf{M}_{T} \odot \mathbf{Y}_{u}^{(2)} + (\textbf{1}-\mathbf{M}_{T}) \odot \mathbf{Y}_{u}^{(1)},
  \end{aligned}
  \end{cases}
\end{equation}
}

\noindent where \textbf{1} is an all-one matrix and $\odot$ represents the element-wise multiplication.

\noindent 3) \textbf{Intra- and Inter-domain Mixing.}
When these two strategies, tailored for scene text images, are integrated within the student-teacher framework, they collectively form a comprehensive method that effectively harnesses the strengths of both synthetic and real data. The whole process can be formulated as:
{\setlength\abovedisplayskip{2pt}
\setlength\belowdisplayskip{2pt}
\begin{equation}
  \begin{cases}
  \begin{aligned}
    &\texttt{A}(\mathbf{X}) \leftarrow \texttt{A}^{G}\big(\texttt{A}^{T}(\mathbf{X}_{u}),\mathbf{X}_{l}\big), \\
    &\texttt{A}(\mathbf{Y}) \leftarrow \texttt{A}^{G}\big(\texttt{A}^{T}(\mathbf{Y}_{u}),\mathbf{Y}_{l}\big)
  \end{aligned}
  \end{cases}
\end{equation}
}

\noindent Consequently, the synergy maximizes the utilization of synthetic labels as a valuable resource, effectively tackling the intricacies of real-world scenarios.

\subsection{Discussion on Character Region Awareness} \label{CRA}
Existing LSD only contains English texts with tightly spaced characters, while URD may encompass text properties from various languages (such as Japanese, Korean, Chinese, \emph{etc}). Consequently, even though the model predictions effectively cover text regions, we encounter challenges in parsing these predictions into word-level bounding boxes or polygons.

Indeed, when using a single language (\emph{e.g.,} English) as the source domain and adapting it to various language domains, 
it appears that no matter how sophisticated the models and loss functions are, they struggle to accurately and effectively parse the complex text layout and semantic context. This challenge primarily arises from variations in the spacing and arrangement of characters, words, and sentences across languages. 
This observation leads us to revisit the learning objective of scene text detectors during pre-training. 

We emphasize an alternative perspective \textcolor{black}{\textemdash} utilizing characters as the fundamental learning unit for text detection. 
The strategy presents several advantages:
1) Like word-level bounding boxes (WBB), character-level bounding box (CBB) annotations can be readily acquired from the existing synthetic dataset (SynthText) at no additional cost;
2) Instead of parsing complex layouts and spacing disparities across languages during pre-training, decoding them into individual characters is straightforward and attainable;
3) The pre-trained model with character-level region awareness easily integrates its learned character features to adapt text attribute rules provided by different language text datasets during fine-tuning.
To this end, we select CBB annotations as $\mathcal{B}_{l}$. 
% Following the pipeline outlined in Fig.\,\ref{fig:3}, the mixed image $\mathbf{X}_{g}$ is created by cropping the synthetic image $\mathbf{X}_{l}$ using CBB annotations $\mathcal{B}_{l}$ and pasting its glyphs onto the unlabeled image $\mathbf{X}_{u}$.
% The corresponding mixed pseudo labels $\mathbf{Y}_{g}$ are then obtained by combining the pixel-wise binary map $\mathbf{Y}_{l}$ which is converted from $\mathcal{B}_{l}$, with the estimated character-level pseudos $\mathbf{Y}_{u}$ generated by the teacher branch.
\section{Experiment}
\subsection{Dataset}
\noindent \textbf{Labeled Synthetic Data (LSD).} The SynthText~\cite{gupta2016synthetic} serves as our source of LSD, consisting of 800K images. Each image is provided with character-level and word-level bounding boxes, along with corresponding text transcriptions. 

\noindent \textbf{Unlabeled Real Data (URD).} URD is collected from a wide array of natural scenes, encompassing a diverse range of fonts, languages, and text shapes.
Specifically, URD includes 430K images from LSVT~\cite{sun2019icdar}, 9K from MLT17~\cite{nayef2017icdar2017}, 19657 from MLT19~\cite{nayef2019icdar2019}, 229 from ICDAR2013~\cite{karatzas2013icdar}, and 500 from USTB-SV1K~\cite{SV1K}. 

\noindent \textbf{Benchmarks.} Four publicly available benchmarks are used to evaluate the per-\par 

\noindent formance, including 1) Total-Text (TT)~\cite{ch2020total}, 2) CTW 1500 (CTW)~\cite{liu2019curved}, 3) ICDAR2015 (IC15)~\cite{karatzas2015icdar}, and 4) MSRA-TD500 (TD)~\cite{yao2012detecting}.

\subsection{Implementation Details}
Referring to recent works~\cite{SKTM,VLPT,oCLIP}, we combine representative text detection methods~\cite{DB,PAN,PSENet,FCENet} to conduct experiments. URD is used in Tabs. \ref{tb:expr1}-\ref{tb:MT} and the same data is used in Tabs. \ref{tb:Model Structure}-\ref{tb:ssl}. All experiments are implemented on a server with 2 NVIDIA A800 GPUs, using PyTorch.
\noindent \textbf{1) Pre-Training.} The pre-training experiments are conducted on LSD and URD, utilizing the open-source codes provided by these text detection methods. For synthetic images, we adopt consistent augmentation settings as prescribed by these methods for a fair comparison. For unlabeled real images, we employ a standard weak-to-strong augmentation technique~\cite{AugSeg} to enrich diversity. 
Our method is trained using the AdamW optimizer~\cite{AdamW}, along with a cosine learning rate scheduler~\cite{Cosine}. This includes a warm-up phrase for 1 epoch in a total of 10 epochs. The base learning rate is set to $0.003 \times \rm{batchsize} / 256$.
At each iteration, we equally sample LSD and URD with a total batch size of 24. 
\noindent \textbf{2) Fine-Tuning.}
For a fair comparison, we employ the same fine-tuning configurations as these detection methods and then evaluate them on these publicly available benchmarks: TT, CTW, IC15 and TD, following the standard metric: Precision (P), Recall (R), and F-measure (F). 

\subsection{Experiment Results}

To ensure a comprehensive comparison, we collect results from public evaluation benchmarks reported by existing pre-training methods, and organize them in Table \ref{tb:expr1}. The sources of these results are clearly recorded in the table's caption using different symbols for easy reference.
Specifically, these methods can be classified into three pre-training paradigms. 
In Paradigm A, models (\emph{i.e., +ST, +TCM}) are initially pre-trained on LSD and then fine-tuned on real datasets directly. In Paradigm B, models (\emph{i.e., +STKM, +VLPT, +oCLIP}) are first pre-trained on LSD by various pretext tasks, and the well-learned backbone weights are transferred into scene text detectors when fine-tuning on real datasets. In Paradigm C, models (\emph{i.e., +MTM}) are pre-trained on URD by pretext tasks and on LSD by supervised learning, and then fine-tuned on real datasets.

\noindent \textbf{1) Results on DBNet.} As indicated in the second group of rows in Table \ref{tb:expr1}, all pre-training methods both report their cooperative performance with DBNet. 
When compared to methods using paradigm A, differing from \emph{DBNet+ST}, we exclude the use of a deformable convolutional network (DCN) in our model.
Despite these differences, our method still achieves impressive results, yielding improvements of 4.2\%, 4.5\%, 4.6\%, and 5.2\% on TT, CTW, IC15, and TD datasets, respectively.
Compared to \emph{DBNet+TCM} methods, our method achieves SOTA detection performance on four datasets, particularly on curved text datasets (TT and CTW), with gains of 3.0\% and 2.8\%, respectively.
In comparison with methods following Paradigm B, our method consistently outperforms previous SOTA works by substantial margins, achieving performance improvements of 2.6\%, 5.9\%, 3.5\% and 1.6\% on TT, CTW, IC15, and TD datasets, respectively. In comparison with \emph{DBNet+MTM} following Paradigm C, our method gets 3.2\% and 4.5\% gains on TT and TD datasets, respectively.

\begin{table}[t]
  \setlength\tabcolsep{4pt}
  \centering
  \caption{Comparison with SOTA pre-training methods for text detectors on TT, CTW, IC15, and TD. F-measure (\%) is reported. 
  $^\divideontimes$ denotes the use of extra labeled real data.
  For easy reference, we mark these symbols $\dagger$,\,$^\ast$,\,$\ddagger$,\, and $^\star$ separately to indicate the results from \cite{FastTCM}, \cite{VLPT}, \cite{oCLIP}, and \cite{MTM}, within the scope of this paper. The best results are shown in \textbf{bold font}. The second-best results are underlined.
  }
  \scalebox{0.8}{
  \begin{tabular}{c|l|c|llll}
      \toprule 
      Index & Method & Venue & TT    & CTW  & IC15  & TD  \\
      \hline 
      \multirow{8}*{1}
      % &SegLink~\cite{SegLink} & CVPR'17            &-     &-     &-     &77.0  \\
      &PSENet-1s~\cite{PSENet} &CVPR'19                            &80.9  &82.2  &85.7  &-     \\
      % &LOMO~\cite{LOMO} &CVPR'19                                   &81.6  &78.4  &87.2  &-     \\
      % &MOST~\cite{MOST} &CVPR'21                                   &-     &-     &88.2  &86.4  \\
      &PAN++~\cite{PAN++} &PAMI'21                                 &85.3  &84.0  &84.5  &87.9  \\
      &DBNet++~\cite{DB++} &PAMI'22                                &86.0  &85.3  &87.3  &87.2  \\
      &DeepSOLO$^\divideontimes$ ~\cite{DeepSolo} &CVPR'23                           &87.6  &-     &89.8     &-     \\
      &ESTextSpotter$^\divideontimes$ ~\cite{ESTextSpotter} &ICCV'23                 &90.0  &-     &91.0  &89.5     \\
      &SIR~\cite{SIR} &ACM MM'23                                                   &88.2  &85.5  &87.8  &89.6     \\
      \hline  
      \multirow{7}*{2}&DBNet+ST$\dagger$~\cite{DB} &AAAI'20           &84.7  &83.4  &85.4  &84.9  \\
      &DBNet+STKM$\dagger$~\cite{SKTM} &CVPR'21                       &85.5  &-     &86.1  &85.9  \\
      &DBNet+VLPT$\dagger$~\cite{VLPT} &CVPR'22                       &\underline{86.3}  &-     &86.5  &88.5  \\
      &DBNet+oCLIP$\dagger$~\cite{oCLIP} &ECCV'22                     &84.1  &82.0  &85.4  &-        \\
      &DBNet+TCM$\dagger$~\cite{TCM} &CVPR'23                         &85.9  &\underline{85.1}  &\underline{89.4}  &\underline{88.8}  \\
      &DBNet+MTM$^\star$~\cite{MTM} &ACM MM'23                            &85.7  &-     &-     &85.6  \\
      % &DBNet+FastTCM$\dagger$~\cite{FastTCM} &arxiv'23                &86.1  &\underline{85.2}  &\underline{89.5}  &\underline{88.9}  \\
      &DBNet+FreeReal (ours)  & -                                                   &$\mathbf{88.9}_{(+2.6)}$  &$\mathbf{87.9}_{(+2.8)}$  &$\textbf{90.0}_{(+0.6)}$  &$\mathbf{90.1}_{(+1.3)}$  \\
      \hline  
      \multirow{3}*{3}&PANet+ST$\dagger$~\cite{PAN} &ICCV'19         &\underline{85.0}  &83.7  &82.9  &84.1  \\
      &PANet+TCM$\dagger$~\cite{TCM} &CVPR'23                        &-     &\underline{84.3}  &\underline{84.6}  &\underline{85.3}  \\
      % &PANet+FastTCM$\dagger$~\cite{FastTCM} &arxiv'23               &-     &\underline{84.5}  &\underline{84.9}  &\underline{85.4}  \\
      % &PAN+MTM~\cite{MTM} &MM'23                                    &85.8  &-     &-     &-  \\
      &PANet+FreeReal (ours)  & -                                                   &$\mathbf{87.4}_{(+2.4)}$  &$\mathbf{88.4}_{(+4.1)}$  &$\mathbf{86.0}_{(+1.4)}$      &$\mathbf{89.9}_{(+4.6)}$  \\
      \hline  
      \multirow{5}*{4}&PSENet+ST$^\ast$~\cite{PSENet} &CVPR'19           &83.9  &81.6  &81.3  &-     \\
      &PSENet+STKM$^\ast$~\cite{SKTM} &CVPR'21                           &84.3  &82.9  &83.7  &-     \\
      &PSENet+VLPT$^\ast$~\cite{VLPT} &CVPR'22                           &\underline{86.1}  &\underline{83.3}  &\underline{84.3}  &-     \\
      &PSENet+oCLIP$\ddagger$~\cite{oCLIP} &ECCV'22                         &85.5  &82.8  &-     &-     \\
      &PSENet+FreeReal (ours)  & -                                                   &$\mathbf{86.8}_{(+0.7)}$      &$\mathbf{86.1}_{(+2.8)}$  &$\mathbf{85.2}_{(+0.9)}$  &\textbf{86.0}      \\
      \hline
      \multirow{4}*{5}&FCENet+ST$\dagger$~\cite{FCENet} &CVPR'21   &\underline{85.8}  &85.5  &86.2  &85.4  \\
      &FCENet+oCLIP$\ddagger$~\cite{oCLIP} &ECCV'22                &-     &85.6  &86.7  &-     \\
      &FCENet+TCM$\dagger$~\cite{TCM} &CVPR'23                     &-     &\underline{85.9}  &\underline{87.1}  &\underline{86.9}  \\
      % &FCENet+FastTCM$\dagger$~\cite{FastTCM} &arxiv'23            &-     &\underline{86.0}  &\underline{87.3}  &\underline{87.1}  \\
      &FCENet+FreeReal (ours)  & -                                                   &$\mathbf{87.9}_{(+2.1)}$      &$\mathbf{87.2}_{(+1.3)}$      &$\mathbf{87.7}_{(+0.6)}$      &$\mathbf{89.0}_{(+2.1)}$      \\  
      \bottomrule
      \end{tabular}
      }
      \label{tb:expr1}
\end{table}

\noindent \textbf{2) Results on PANet.} As shown in the third group of rows in Table \ref{tb:expr1}, our method outperforms \emph{PANet+ST} on four benchmarks, achieving a gain of 3.85\% in terms of average F-measure, while using the same parameters and labeled data for pre-training and fine-tuning. 
Furthermore, our method exhibits its prominent superiority, particularly on the CTW and TD datasets, surpassing the \emph{PANet+TCM} method by a substantial margin of 4.1\% and 4.6\%, respectively.

\noindent \textbf{3) Results on PSENet.} FreeReal brings 0.7\%, 2.8\%, and 0.9\% gains on TT, CTW, and IC15 datasets, respectively. 

\noindent \textbf{4) Results on FCENet.} 
FreeReal further refreshes the best text detection results and improves 2.1\%, 1.3\%, 0.6\% and 2.1\% on TT, CTW, IC15, and TD datasets, respectively. These experimental results demonstrate the robustness of FreeReal, even when cooperated with regression-based methods.

\begin{table}[t]
  \setlength\tabcolsep{4pt}
  \centering
  \caption{Comparison with existing pre-training paradigms when fine-tuning on the TotalText dataset.
  These symbols $\dagger$,\,$\ddagger$,\, and $^\star$ indicate the results from \cite{FastTCM}, \cite{oCLIP} and \cite{MTM}, respectively. $\Delta$F is the improvement relative to the baseline.}
  \scalebox{0.8}{
  \begin{tabular}{c|l|c|cc|c|c}
      \toprule 
      \multirow{2}{*}{Paradigm}&\multirow{2}{*}{Method} &Venue & \multicolumn{2}{c|}{Data} &\multirow{2}{*}{F-measure} & \multirow{2}{*}{$\Delta$ F(\%)}\\
              &       & &LSD &URD                  &   &\\
      \hline
      -&DBNet$\ddagger$~\cite{DB} &AAAI'20 &- &- &78.2 & -\\
      \hline  
      \multirow{2}*{A}&DBNet+ST$\dagger$~\cite{DB}  &AAAI'20 &800K &- &84.7 & +6.5\%\\
      &DBNet+TCM$\dagger$~\cite{TCM}                &CVPR'23 &800K &- &85.9 & +7.7\% \\
      % &DBNet+FastTCM$\dagger$~\cite{FastTCM}        &800K &- &86.1 & +7.9\%  \\
      \hline
      \multirow{3}*{B}&DBNet+STKM$\dagger$~\cite{SKTM}  &CVPR'21 &800K &- &85.5 &+7.3\% \\
      &DBNet+VLPT$\dagger$~\cite{VLPT}                  &CVPR'22 &800K &- &\underline{86.3} &+8.1\% \\
      &DBNet+oCLIP$\ddagger$~\cite{oCLIP}                &ECCV'22 &800K &- &84.1 &+5.9\% \\
      \hline
      \multirow{1}*{C}&DBNet+MTM$^\star$~\cite{MTM}                    &ACM MM'23 &800K &41K &85.7 &+7.5\% \\
      \hline 
      \multirow{5}*{D}&Ours (9\%)             &- &71K  &41K	&\textbf{87.4}	&\textbf{+9.2\%} \\
      &Ours (10\%)             &- &80K  &46K	&\textbf{87.5}	&\textbf{+9.3\%} \\
      &Ours (20\%)                             &- &160K &92K	&\textbf{88.1}	&\textbf{+9.9\%} \\
      &Ours (50\%)                             &- &400K &230K&\textbf{88.3}	&\textbf{+10.1\%} \\
      &Ours (100\%)                            &- &800K &460K&\textbf{88.9}	&\textbf{+10.7\%} \\
      \bottomrule      
      \end{tabular}
      }
      \label{tb:DF}
\end{table}
\noindent \textbf{5) Paradigm Comparison.}
In this section, we aim to investigate which kinds of pre-training paradigms for text detectors are important and meaningful for future studies.
To this end, we conduct a comparative analysis with three mainstream pre-training paradigms mentioned above, as summarized in Table \ref{tb:DF}.
When compared to these well-curated designs at the cost of introducing additional auxiliary pretext tasks or more training modules, our proposed paradigm achieves a significant improvement of 1.2\% (87.5 vs. 86.3), despite utilizing 10\% of the labeled synthetic data and 10\% of the unlabeled real data. This outcome underscores our ability to harness the potential of unlabeled real images. 
Moreover, when we gradually increase the proportion of the total training data, the text detection performance continues to improve. Specifically, our method achieves gains of 1.8\%, 2.0\%, and 2.6\% at data fractions of 20\%, 50\%, and 100\%, respectively, compared to the recent SOTA method. 
This outcome fully demonstrates the superiority of our pre-training paradigm, which jointly leverage LSD and URD by bridging their gap for enhancing the text detector pre-training.

\noindent \textbf{7) Results on Unseen Language Text.} 
Our pre-trained model is capable of learning the basic character feature representations of languages and then combining them during fine-tuning. To show its robustness, we choose to evaluate on unseen language text dataset VinText~\cite{VinText}. FreeReal achieves an improvement of 0.7\%, as shown in Table \ref{tb:MT}.
\section{Ablations and analysis}
Unless otherwise stated, all ablation experiments are implemented in collaboration with DBNet method~\cite{DB} using 20\% of LSD and 20\% of URD and evaluated\par 
\noindent on the curved TotalText (TT) dataset for efficiency.

\noindent \textbf{Effectiveness of Network Components.}
To comprehensively evaluate the effectiveness of our proposed method, we divide it into three components and evaluate each one individually, as shown in Table \ref{tb:Model Structure}. ``Stu-Tea'', ``CRA'', ``GlyphMix'' denote the standard student-to-teacher framework, Character Region Awareness introduced in Sec.\ref{CRA} to bridge the linguistic domain gap, and the glyph-based mixing mechanism, respectively.
First, we enhance the baseline DBNet model by incorporating a standard student-teacher framework, yielding a performance gain of 1.31\%.
Second, within the student-teacher framework, we evaluate the individual contributions of the CRA and GlyphMix components, achieving 1.73\% and 3.12\% performance improvements, respectively. Finally, upon combining all plug-ins, the overall performance reaches 88.1\%, showing a substantial gain of 3.40\% compared to the baseline. 
These experiments consistently demonstrate that FreeReal effectively bridges both the synth-to-real and language-to-language domain gaps when leveraging the intrinsic qualities of unlabeled real data. 

\begin{table}[t]
  \setlength\tabcolsep{4pt}
  \centering
  \caption{Comparison with supervised SOTA work on unseen language text. ``R50" and ``DeT" denote a ResNet50 and Deformable Transformer structure, respectively. 
  $^\divideontimes$ denotes the use of extra labeled real data.}
  \scalebox{0.8}{
  \begin{tabular}{l|c|c|c|c|c}
      \toprule 
      Method &Venue&Backbone &Para. &FPS &F\\
      \hline
      ESTextSpotter$^\divideontimes$ &ICCV'23&R50+DeT&56M&3.4 &89.6\\
      DBNet+FreeReal (ours)&-&R50&\textbf{25M}&\textbf{11}&\textbf{90.3}\\
      \bottomrule      
      \end{tabular}
      }
      \label{tb:MT}
\end{table}

\newcolumntype{L}[1]{>{\raggedright\arraybackslash}p{#1}}
\newcolumntype{C}[1]{>{\centering\arraybackslash}p{#1}}
\newcolumntype{R}[1]{>{\raggedleft\arraybackslash}p{#1}}
\begin{table}[t]
  \setlength\tabcolsep{4pt}
  \centering
  \caption{Evaluate the effectiveness of the proposed modules. $\Delta$F is the improvement of F-measure relative to baseline.}
  \scalebox{0.85}{
  \begin{tabular}{C{14mm}C{10mm}c|c|C{7mm}C{7mm}C{7mm}}
    \toprule 
      Stu-Tea   & CRA        & GlyphMix       & $\Delta$ F(\%) &P &R &F\\
    \hline
      $\times$   & $\times$   & $\times$  & - &87.1  &82.5  & 84.7\\
      \checkmark & $\times$   & $\times$  & +1.31\%  &88.80 &83.39 & 86.01\\
      \checkmark &\checkmark  &$\times$   & +1.73\%  &87.76 &85.15 &86.43  \\
      \checkmark &$\times$   &\checkmark & +3.12\%  &89.10 &86.58  &87.82  \\
    \hline
      \checkmark &\checkmark &\checkmark & \textbf{+3.40\%}  &88.81	&87.40	& \textbf{88.10} \\
    \bottomrule 
    \end{tabular}
}
\label{tb:Model Structure}
\end{table}

\newcolumntype{L}[1]{>{\raggedright\arraybackslash}p{#1}}
\newcolumntype{C}[1]{>{\centering\arraybackslash}p{#1}}
\newcolumntype{R}[1]{>{\raggedleft\arraybackslash}p{#1}}
\begin{table}[t]
  \setlength\tabcolsep{4pt}
  \centering
  \caption{Evaluate the effectiveness of GlyphMix components. ``GIM" and ``TIM" denote the Graffiti-like Inter-domain Mixing and Text-balanced Intra-domain Mixing.}
  \scalebox{0.9}{
  \begin{tabular}{C{10mm}C{10mm}|c|cC{7mm}C{7mm}C{7mm}}
    \toprule 
      GIM & TIM  & $\Delta$ F(\%) &P &R &F\\
    \hline
      $\times$  &$\times$   & -  &87.76 &85.15 &86.43  \\
      $\times$ &\checkmark   & +1.22\%  &88.64	&86.68	&87.65\\
      \checkmark &$\times$ & +1.28\%  &88.29	&87.13	&87.71\\
    \bottomrule 
    \end{tabular}
}
\label{tb:GlyphMix Structure}
\end{table}

\noindent \textbf{Discussion on Using WBB or CBB.}
The SynthText dataset offers word-level and character-level annotations, with the latter being exploited in our method to bridge the linguistic domain gap, bringing a 0.4\% gain as shown in the third row of Table \ref{tb:Model Structure}. Notably, even without the use of CRA, our method still achieves an F-measure of 87.82\% (in the fourth row), significantly surpassing recent SOTAs.

\noindent \textbf{Comparison with pre-training methods using URD.} Bridging the synth-to-real gap to simultaneously leverage LSD and URD for pre-training is our major contribution. This capability, not present in other PSTD methods, enables seamless incorporation of the huge number of URD to enhance pre-training.
Thus to compare to other PSTD methods with complex pretext task designs, we use different ratios of URD to explore the effectiveness of our pre-training paradigm. Even when using 10\% LSD and 10\% URD in Table \ref{tb:DF}, our method still achieves SOTA with a 1.2\% gain, which exhibits our method can \emph{serve as a novel baseline for PSTD}. Besides, we add the comparison \emph{using additional URD to train other PSTD methods.} In this case, these mentioned methods can be viewed the same as MTM~\cite{MTM}. As a result, FreeReal outperforms MTM by 1.7\% in Table \ref{tb:DF}.

\noindent \textbf{Effectiveness of GlyphMix Components.} In Table \ref{tb:GlyphMix Structure}, GlyphMix comprises two modules, inter-domain GIM and intra-domain TIM, contributing gains of 1.28\% and 1.22\% to the final performance, respectively.
Besides, when compared to the data engine tool~\cite{gupta2016synthetic}, GIM achieves intra-domain mixing in a cost-effective way, which yields real-world images online via K-means, a more flexible, faster and direct process than the multi-stage, pre-generated method. 
Specifically, GIM attains a 10.7x speedup (2.6h \emph{vs} 27.7h) in generating 92K images, which is more suitable for real-time online training tasks.
Additionally, GIM can avoid improper text blends on URD by leveraging the learned knowledge (boxes/maps) from our pre-training framework online. We also conduct the ablations and get 0.5\% (85.3\% \emph{vs} 84.8\%) performance gains, when using 5\% training data.

\begin{table}[h]
  \setlength\tabcolsep{4pt}
  \centering
  \caption{Comparison of the effectiveness with existing domain bridging methods. ``DCA'' refers to \textbf{D}omain \textbf{C}lassification \textbf{A}ccuracy in aligning with the real domain.}
  \scalebox{0.85}{
  \begin{tabular}{L{25mm}|c|ccc|c}
      \toprule 
      method & DCA &P &R &F & $\Delta$ F(\%)\\
      \hline  
      Stu-Tea  &- &88.80 &83.39 & 86.01 & -\\
      Mixup\cite{mixup}   &69.5 &89.17 &84.12 &86.57                & +0.56\%  \\
      CutMix\cite{cutmix}  &72.8 &87.99 &85.64	&86.80  & +0.79\% \\
      ClassMix\cite{classmix} &\underline{78.9} &90.65 &83.61 &\underline{86.99}  & +0.98\% \\
      \hline 
      FreeReal (ours)    &\textbf{95.7} &88.81	&87.40	&\textbf{88.10} & \textbf{+2.09\%} \\
      \bottomrule      
      \end{tabular}
      }
      \label{tb:Domain}
\end{table}
\begin{table}[!t]
  \setlength\tabcolsep{4pt}
  \centering
  \caption{Comparison of the effectiveness with SSL methods.}
  \scalebox{0.85}{
  \begin{tabular}{c|c|ccc|c}
      \toprule 
      Method & Venue &P &R &F & $\Delta$ F(\%)\\
      \hline  
      Stu-Tea & -  &88.80 &83.39 & 86.01 & -\\
      GTA\cite{GTA} &NeurIPS'22   &87.18 &81.08 &84.02 & -1.99\%\\
      BCP\cite{BCP} &CVPR'23  &86.72	&85.82 &86.27 & +0.26\%\\
      AugSeg\cite{AugSeg} &CVPR'23   &88.48 &85.65 &87.04 & +1.03\%\\
      Unimatch\cite{Unimatch} &CVPR'23  &88.94 &85.56 &87.22 & +1.21\%\\
      \hline 
      FreeReal (ours)  &-    &88.81	&87.40	&\textbf{88.10} & \textbf{+2.09\%} \\
      \bottomrule      
      \end{tabular}
      }
      \label{tb:ssl}
\end{table}

\noindent \textbf{Comparison with Bridging-domain Methods.} The goal of generating mixed images through domain-bridging methods is to bring synthetic images closer to real images. Consequently, the adaptation capability can be viewed as the likelihood of mixed images belonging to the real domain. A classification task is employed to assess the capability, whose the evaluation protocol is defined: \emph{Given a well-learned domain classifier $\phi$ (trained by n synthetic images $x_{l}\in X_{L}$ and n real images $x_{u}\in X_{U}$) and domain bridging method $\xi$, the classifier predicts the domain label (0 for synthetic and 1 for real) of mixed images and evaluates the accuracy as $\frac{\sum_{x_{l}\in X_{L},x_{u}\in X_{U}}\phi(\xi(x_{l}, x_{u}))}{n^{2}}$.} To accomplish this, we employ a network architecture, consisting of resnet18 as a feature extractor, followed by a binary classification head, for simplicity. Subsequently, the performance of different domain bridging methods is summarized in Table \ref{tb:Domain}. Remarkably, when compared to these methods, GlyphMix achieves an outstanding performance score of 95.7\%, surpassing the second-best method by a substantial margin of 16.8\%. 
This result demonstrates that our method is able to generate arbitrarily realistic images without domain drift.

To be more convincing, we also conduct text detection experiments to assess the effectiveness. As depicted in Table \ref{tb:Domain}, our method consistently outperforms others, achieving the best F-measure value. Furthermore, the observed positive correlation between DCA and F-measure reinforces the idea that bridging the domain gap is the crucial factor in improving text detection performance.

\noindent \textbf{Comparison with Semi-supervised Methods.}
In this section, we compared the performance with recent SSL methods. These methods tend to adopt increasingly complex designs, involving advanced contrastive learning techniques, additional network components, multiple ensemble models, and complex training procedures. 
In contrast, we aim to eliminate domain gaps between LSD and URD. Despite adopting the standard student-teacher framework, our method still exceeds the previous SOTA by 0.88\% in Table \ref{tb:ssl}.

\noindent \textbf{Effectiveness of $\mathbf{M}_{G}$.}
We evaluate the accuracy of character segmentation results $\mathbf{M}_{G}$ using TotalText~\cite{ch2020total} and TextSeg~\cite{TextSeg} datasets, both of which provide pixel-level annotations. 
Specifically, compared to conducting the cluster task on the whole image, we crop it into several text instances and then perform clustering to form $\mathbf{M}_{G}$. While the accuracy of the former is 32.35\% for TotalText and 45.54\% for TextSeg, $\mathbf{M}_{G}$ achieves 80.43\% and 84.14\%, respectively. Notably, the underlying morphological representations of glyphs exhibit relative invariance to slight structural changes, such as thicker or thinner. This reduces the dependence on pixel-level high-precision segmentation with expensive costs.
Furthermore, although the clustering process introduces some noise, leading to inaccuracies in $\mathbf{M}_{G}$, it can still be viewed as a way of adding data noise in the image-mixing process, enhancing the robustness of text detection. 

\noindent \textbf{Ablation Study on Hyper-parameters.}
We study the impact of the parameter $\gamma$ on the pseudos generated by the teacher model during pre-training. This parameter controls the proportion of unreliable and reliable pixels. 
With $\gamma$ set at 80, 50, and 20, the F-measure are 87.94\%, 87.68\%, and 87.06\%, respectively. Finally, we employ a linear warm-up strategy for $\gamma$ that gradually reduces from 80 to 20, yielding the best result with an F-measure of 88.10\%.
\section{Conclusion}
We propose an effective pre-training paradigm for scene text detection, situated within a standard student-teacher framework. The paradigm leverages the complementary strengths of both labeled synthetic data (LSD) and unlabeled real data (URD), standing out from existing intricate pre-training paradigms in text detection due to its simplicity and significant advancements.
Moreover, we identify and address prominent challenges, \emph{i.e.,} synth-to-real and language-to-language domain gaps. The proposed solutions, namely GlyphMix and character region awareness, have shown their effectiveness in mitigating these challenges and improving text detection performance, as substantiated by extensive experimental results. 
More importantly, our method exhibits the ability to harness the intrinsic attributes of URD, achieving substantial performance improvements even with a small fraction of LSD. This highlights the significance of exploring the potential of unlabeled real data in scene text detection, ultimately advancing the field and its practical applications.

\noindent \textbf{Acknowledgment.} 
This work was supported by the NSFC under Grant 62322604 and 62176159, and in part by the Shanghai Municipal Science, Technology Major Project under Grant 2021SHZDZX0102, and National Key R\&DProgram of China (NO. 2022ZD0160100).

\par\vfill\par

% ---- Bibliography ----
%
% BibTeX users should specify bibliography style 'splncs04'.
% References will then be sorted and formatted in the correct style.
%
\bibliographystyle{splncs04}
\bibliography{main}
\end{document}